\documentclass[10pt,journal,compsoc]{IEEEtran}
\usepackage{hyperref}
\usepackage{graphicx}
\usepackage[flushleft]{threeparttable}
\usepackage{booktabs} 
\usepackage[caption=false]{subfig}
\usepackage{amsfonts}
\usepackage{multirow}
\usepackage{tabularx}
\usepackage{xcolor}

\newcolumntype{M}{>{$}c<{$}}
\newcolumntype{C}{>{\centering\arraybackslash}p}
\newcolumntype{Y}{>{\centering\arraybackslash}X}

\ifCLASSOPTIONcompsoc
  \usepackage[nocompress]{cite}
\else
  \usepackage{cite}
\fi
\ifCLASSINFOpdf
\else
\fi
\begin{document}
%
\title{PrintsGAN: Synthetic Fingerprint Generator}
%
%
%
%

\author{Joshua~J.~Engelsma,~\IEEEmembership{Member,~IEEE}, Steven~A.~Grosz, and~Anil~K.~Jain,~\IEEEmembership{Life~Fellow,~IEEE}
\IEEEcompsocitemizethanks{\IEEEcompsocthanksitem J.J. Engelsma, S.A. Grosz and A.K. Jain are with the Department of Computer Science and Engineering, Michigan State University, East Lansing, MI, 48824 USA (e-mail: engelsm7@cse.msu.edu, groszste@cse.msu.edu, jain@cse.msu.edu). J.J. Engelsma is now with Amazon but contributed to this work while finishing up his PhD at Michigan State University.}
}

\markboth{Journal of \LaTeX\ Class Files,~Vol.~14, No.~8, August~2015}%
{Engelsma \MakeLowercase{\textit{et al.}}: PrintsGAN: Synthetic Fingerprint Generator}
%



\IEEEtitleabstractindextext{%
\begin{abstract}
A major impediment to researchers working in the area of fingerprint recognition is the lack of publicly available, large-scale, fingerprint datasets. The publicly available datasets that do exist contain very few identities and impressions per finger. This limits research on a number of topics, including e.g., using deep networks to learn fixed length fingerprint embeddings. Therefore, we propose PrintsGAN, a synthetic fingerprint generator capable of generating unique fingerprints along with multiple impressions for a given fingerprint. Using PrintsGAN, we synthesize a database of 525K fingerprints (35K distinct fingers, each with 15 impressions). Next, we show the utility of the PrintsGAN generated dataset by training a deep network to extract a fixed-length embedding from a fingerprint. In particular, an embedding model trained on our synthetic fingerprints and fine-tuned on a small number of publicly available real fingerprints (25K prints from NIST SD302) obtains a TAR of 87.03\% @ FAR=0.01\% on the NIST SD4 database (a boost from TAR=73.37\% when only trained on NIST SD302). Prevailing synthetic fingerprint generation methods do not enable such performance gains due to i) lack of realism or ii) inability to generate multiple impressions per finger. We plan to release our database of synthetic fingerprints to the public.
\end{abstract}

}

\maketitle

\IEEEdisplaynontitleabstractindextext

%
\IEEEpeerreviewmaketitle

\IEEEraisesectionheading{\section{Introduction}\label{sec:introduction}}

%
%
%
%
\IEEEPARstart{O}{ver} the past several decades, automated fingerprint recognition systems have proliferated into many different facets of our day to day lives including mobile authentication and payments, border crossings and immigration, and various access control terminals~\cite{handbook}. Although fingerprint recognition technology has significantly matured in recent years (now obtaining a False Non-Match Rate of only $0.626\%$ at a False Match Rate of $0.01\%$ on the FVC-ongoing 1:1 hard benchmark~\cite{fvc_ongoing}), there remain unsolved problems which need to be addressed.  One of the main obstacles preventing researchers from adequately addressing these problems is the lack of publicly available fingerprint datasets. In particular, large-scale datasets of fingerprints (with many fingers and multiple fingerprint impressions per finger) are a necessity for i) training the parameters of the various algorithms in the fingerprint recognition pipeline and ii) evaluating the efficacy and speed of the respective algorithms. 

\begin{figure}[!t]
\includegraphics[scale=1.0]{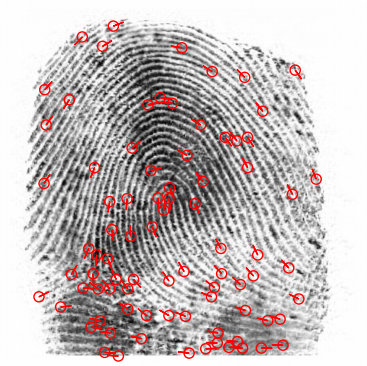} 
\caption{Example of a rolled fingerprint synthesized by PrintsGAN and overlaid with its minutiae representation. Minutiae are automatically annotated with the Verifinger v12 SDK. The fingerprint here qualitatively shows the realism of the fingerprints generated by PrintsGAN.}
\label{fig:fig1}
\end{figure}

For example, in~\cite{deepprint}, the authors used a subset of $455K$ fingerprints of $38K$ fingers from a \textit{privately held}, operational, forensic database to train a deep network, called DeepPrint, to extract highly discriminative fixed-length (192D) fingerprint representations, or embeddings. In contrast to the prevailing variable length, unordered minutiae representation (Fig.~\ref{fig:fig1}), the DeepPrint representation can be matched at orders of magnitude faster speed (useful for large-scale search) and can be matched in the encrypted domain (using a fully homomorphic encryption scheme) in a timely manner and with minuscule loss of accuracy~\cite{engelsma2020hers}. To date, relatively few works~\cite{index1, index2, index3, index4} have pursued developing deep networks, like DeepPrint, to extract fixed-length fingerprint representations, despite the incredible promise of these networks to speed up large scale, accurate fingerprint search and encrypted fingerprint matching~\cite{deepprint}. We posit that the main reason for this is the lack of publicly available fingerprint data (similar to the privately-held forensic dataset used in~\cite{deepprint}) to train such models. Note, nearly all state-of-the-art \textit{face} recognition systems are now extracting deep face representations due in large part to the plethora of face data which has been historically, easily downloaded and aggregated for free from the internet and subsequently used to adequately train deep face networks~\footnote{Presently, even these face datasets are under considerable criticism for violating user privacy as defined in GDPR and other regulations which prohibit use of biometric data without user consent. As a result, many of the previously available face recognition datasets are no longer available for download. This has prompted efforts to generate synthetic faces, see e.g., \cite{shoshan2021gan, karras2017progressive, karras2019style}.}.

\begin{figure*}[t]
\includegraphics[scale=3.0]{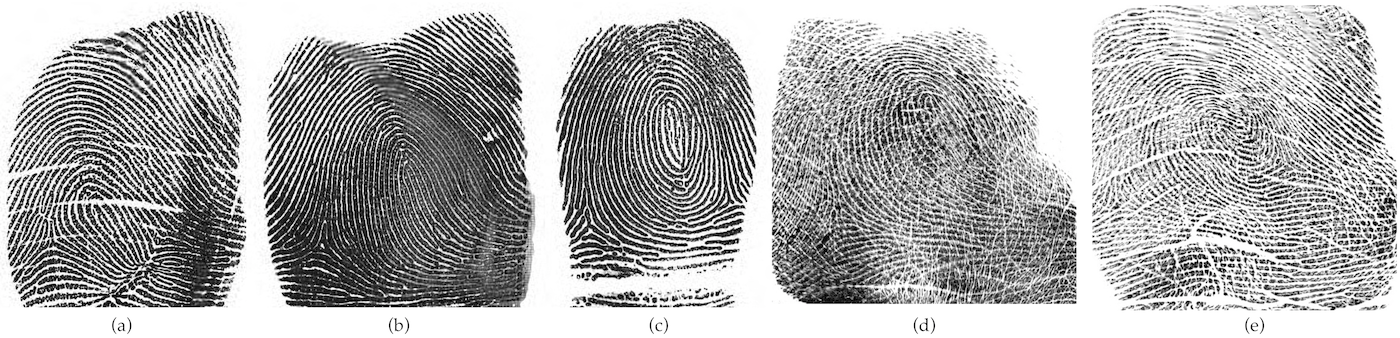} 
\caption{Examples of real fingerprints taken from an operational forensic database (a, b, c)~\cite{longitudinal} and the publicly available NIST SD302 database (d, e)~\cite{nist302}. These fingerprints provide a reference point for qualitatively determining the realism of the synthetic fingerprints shown throughout the paper.}
\label{fig:real_fingerprints}
\end{figure*}

In addition to being limited in algorithm development and training by the lack of publicly available fingerprint training data, researchers are also unable to properly evaluate their algorithms, particularly their large-scale search capability (retrieval accuracy and speed with million or billion scale backgrounds or distractors). Validating the performance of fingerprint search algorithms on large-scale galleries is of significant importance given the integration of fingerprint search algorithms into several real world applications including (i) India's Aadhaar (gallery of $\approx$ 1.3 billion ten-prints~\footnote{\url{https://uidai.gov.in/aadhaar_dashboard/india.php}}) and (ii) the FBI's Next Generation Identification system (NGI)
(gallery of 145.3 million ten-prints~\footnote{\url{https://www.fbi.gov/file-repository/ngi-monthly-fact-sheet/view}}).

While some well-known, publicly available fingerprint data is accessible including the FVC datasets~\cite{fvc2002, fvc2004}, the LivDet datasets~\cite{livdet2019}, and the NIST N2N dataset (NIST SD 302~\cite{nist302}), they are limited in the following ways:

\begin{itemize}
    \item The datasets have a limited number of unique identities (fingers). Of the datasets available, the largest (N2N dataset) has fingerprints from only $2,000$ unique fingers. 
    \item There is a limited number of impressions per identity (\textit{e.g} only 5-10 impressions per finger).
    \item There is no guarantee that the data will remain available to academic researchers. In fact, the widely utilized NIST SD4~\cite{sd4}, NIST SD14~\cite{sd14}, and NIST SD27~\cite{sd27} datasets have all been removed by NIST from their website due to privacy regulations.  
\end{itemize} 

\begin{figure*}[t]
\includegraphics[scale=3.0]{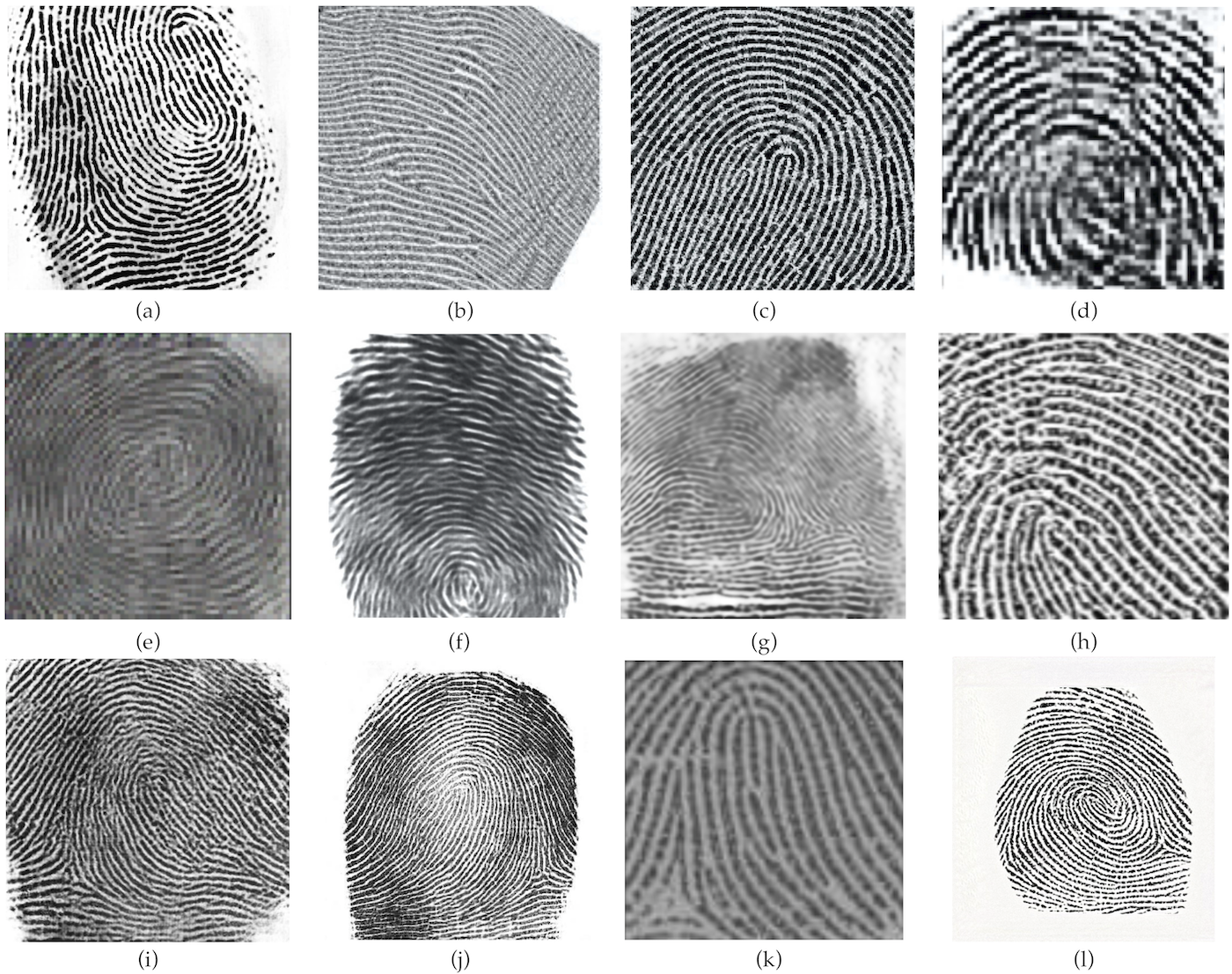} 
\caption{Example images taken from prior fingerprint synthesis algorithms; (a)~\cite{cappelli2002synthetic}, (b)~\cite{zhao2012fingerprint}, (c)~\cite{johnson2013texture}, (d)~\cite{bontrager2018deepmasterprints}, (e)~\cite{finger-gan}, (f)~\cite{attia2019fingerprint}, (g)~\cite{synfi}, (h)~\cite{lightweight}, (i)~\cite{kai_synthetic}, (j)~\cite{mistry2019fingerprint}, (k)~\cite{level-3}, (l)~\cite{bahmani2021high}. Existing synthesis algorithms are limited by a lack of realism (domain gap between real and synthetic fingerprints), \textit{e.g.}, (a-i). GAN based synthesis methods generate more realistic fingerprints \textit{e.g.}, (j-l), however, they are not able to generate multiple impressions for a given fingerprint (they only generate unique fingerprints). Our proposed PrintsGAN generates more realistic fingerprints than the baselines (via a crowd-source evaluation) and is also capable of generating multiple impressions per finger. This enables us to train a CNN on top of our synthetically generated fingerprints to learn a discriminative fingerprint representation for fingerprint matching.}
\label{fig:comparison}
\end{figure*}

\newcommand{\specialcell}[2][c]{%
  \begin{tabular}[#1]{@{}c@{}}#2\end{tabular}}
  \newcommand{\tabitem}{~~\llap{\textbullet}~~}

\begin{table}[t]
\caption{Examples of publicly available fingerprint datasets.}
 \centering
\begin{threeparttable}
\begin{tabular}{c c c c}
 \toprule
 Dataset & \specialcell{FVC 2002 \\DB1 A\tnote{1}} & LivDet 2019\tnote{2} & \specialcell{NIST SD302 \\(N2N)\tnote{3}}\\
 \midrule
 \specialcell{Unique \\Fingers} & 100 & N.A. & 2,000 \\
 \midrule
 \specialcell{Total \\Fingerprints} & 800 & 6,029 & 25,093 \\
 \bottomrule
\end{tabular}
\begin{tablenotes}
\item[1] Other FVC datasets are of similar size.
\item[2] Earlier LivDet datasets are smaller in size.
\item[3] The count of fingerprints in the N2N dataset after cleaning latent / palm slaps and aggregating the data from the individual fingerprint readers.
\end{tablenotes}
\end{threeparttable}
\label{table:fingerprint}
\end{table}

\begin{table}[t]
\caption{Examples of publicly available face datasets.}
 \centering
\begin{threeparttable}
\begin{tabular}{c c c c}
 \toprule
 Dataset & WebFace260M~\cite{zhu2021webface260m} & MS-Celeb~\cite{msceleb} & VGGFace2~\cite{vgg_face}\\
 \midrule
 \specialcell{Number of \\Identities} & 4 Million & 100K & 9,131 \\
  \midrule
 \specialcell{Total \\ Face Images} & 260 Million & 10 Million & 3.3 Million \\
 \bottomrule
\end{tabular}
\end{threeparttable}
\label{table:face}
\end{table}

Tables~\ref{table:fingerprint} and~\ref{table:face} demonstrate the large divide that stands between the amount of publicly available fingerprint data and face data, respectively. In addition to the millions of face images enumerated in Table~\ref{table:face}, the authors in~\cite{charles} demonstrated the ability to develop a web-crawler to download 80 million face images to benchmark the search performance of automated face matchers at scale. This vast quantity of accessible face data has opened up a plethora of promising research directions within the face recognition community and that lamentably remain elusive to fingerprint recognition researchers.

To address the lack of publicly available fingerprint data, numerous studies have been published which describe algorithms to generate synthetic fingerprint images. These fingerprints purportedly do not belong to any real person, and therefore, they do not come attached with stringent Institutional Review Board (IRB) and other privacy regulations~\footnote{Some studies~\cite{tinsley2021face, feng2021gans, nagarajan2018theoretical} suggest that information from the training set may be leaked by Generative Adversarial Networks. In this study, we match our synthetically generated fingerprints against our training set of real fingerprints to verify that no biometric identities from our training set of real fingers are inadvertently leaked.}. However, to date, synthetic fingerprint generators~\cite{cappelli2002synthetic, zhao2012fingerprint, johnson2013texture, bontrager2018deepmasterprints, finger-gan, attia2019fingerprint, synfi, lightweight, kai_synthetic, mistry2019fingerprint, level-3, bahmani2021high} continue to have limited utility for training and evaluating algorithms for the following main reasons:

\begin{itemize}
    \item The fingerprints lack realism. Qualitatively speaking, a human observer can easily differentiate between a synthetic fingerprint and a real fingerprint. In other words there exists a large domain gap between real and synthetic fingerprints (see Fig~\ref{fig:comparison}).
    \item Many of the approaches~\cite{kai_synthetic, mistry2019fingerprint, level-3, bahmani2021high} which attempt to improve the realism of the synthetic fingerprints via advancements in Generative Adversarial Networks (GAN) cannot generate multiple impressions for a given finger or identity. They can only generate unique fingerprint impressions and do not model intra-class variations for a given finger.
\end{itemize}

To address these limitations inherent to prevailing synthetic fingerprint generators, we propose \textit{PrintsGAN}. PrintsGAN utilizes multiple generative adversarial networks (GANs) combined with a style transfer and warping module to generate highly realistic fingerprints. Furthermore, PrintsGAN is able to generate a large variety (distortion, moisture, pressure) of impressions for a given finger. We show qualitatively via a crowd-source evaluation (amongst researchers within the field of fingerprint recognition) that synthetic fingerprints from PrintsGAN are much more similar to real fingerprints than synthetic fingerprints from current methods. We also show this more quantitatively through i) the distribution of minutiae of our synthetic fingerprints compared to real fingerprints, ii) match scores from two state-of-the-art fingerprint matchers (Verifinger v12 SDK and DeepPrint), and iii) NFIQ 2.0 quality scores~\cite{nfiq}.

After demonstrating the realism of the PrintsGAN synthetic fingerprints, we show how the synthetic data from PrintsGAN can be used to train a deep network to extract a fixed-length fingerprint representation useful for improving the search speed against large-scale galleries. In particular, we show that by initializing a deep network with a database of $525,000$ PrintsGAN fingerprints ($35,000$ fingers, $15$ impressions per finger) and then fine-tuning on the publicly available NIST SD302 database ($2,000$ fingers, approximately $12$ impressions per finger) we can obtain a True Accept Rate (TAR) on several real fingerprint evaluation datasets which is substantially higher than if we had trained on just real data (NIST SD302) alone or when pretraining on synthetic fingerprints generated by the previous baseline methods. By showing the ability to train a network on synthetic fingerprints and then perform well on real fingerprints in several additional datasets, we demonstrate that our synthetic fingerprints model the intra-class and inter-class variations of real fingerprints much better than existing synthetic fingerprint generation methods. We also open the door for an entire new research direction within the fingerprint recognition community, namely, to leverage our synthetic data to learn highly discriminative fixed-length fingerprint representations similar to DeepPrint in~\cite{deepprint}.

More concisely, the contributions of this research are as follows:

\begin{itemize}
    \item A synthetic fingerprint generator capable of generating significantly more realistic fingerprints than state-of-the-art methods. We demonstrate this via a crowdsourced study and several quantitative metrics.
    \item The learning of a discriminative fixed-length representation using our synthetic fingerprints. More specifically, we demonstrate that a deep network like~\cite{deepprint} trained on our synthetic fingerprints can be used to search against real fingerprints. Existing synthetic fingerprint generators do not adequately model the inter-class variations and intra-class variations needed to enable learning such discriminative representations.
    \item The creation of a benchmark for learning discriminative fixed-length fingerprint representations using our synthetic fingerprints as training data. To this end, we will release a database of $35$k synthetic fingerprint identities with $15$ impressions each generated from PrintsGAN.
    \item Matching experiments demonstrating that no identity information is ``leaked" from the training database of our synthetic fingerprint generator. This enables us to safely share our synthetic fingerprints with interested researchers to pursue new avenues which were previously inhibited by lack of large-scale public fingerprint datasets.
\end{itemize}

\begin{figure*}[!t]
\includegraphics[scale=1.5]{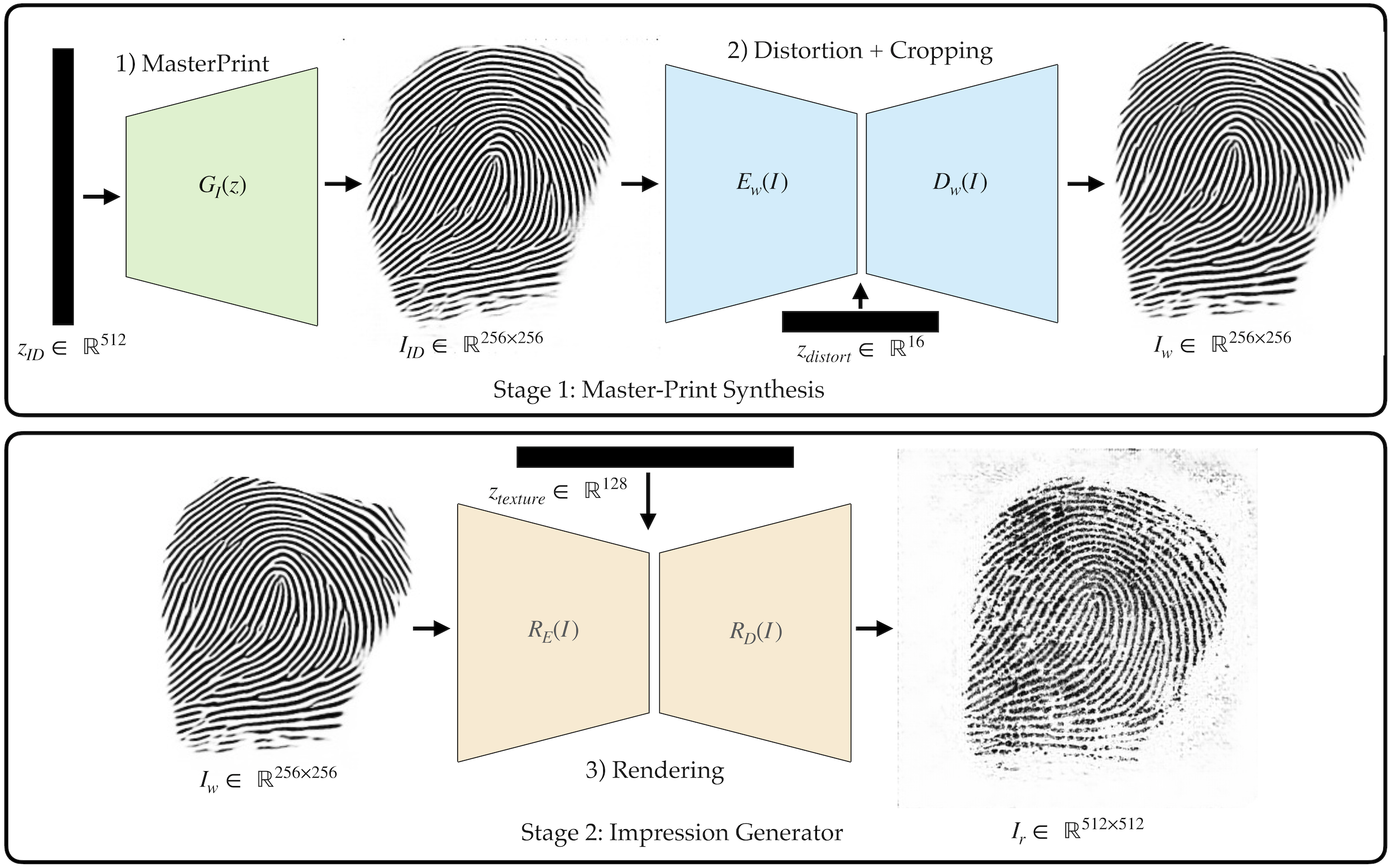} 
\caption{Schematic of PrintsGAN. PrintsGAN operates in two stages. In the first stage, a Master-Print, or a new identity is generated. A Master-Print is a binarized friction ridge pattern at 250 ppi. After synthesizing a Master-Print, it is passed to a non-linear warping and cropping module to simulate the effects of pressing the finger against a fingerprint reader platen at different roll, pitch, yaw, and degree of pressure. Finally, this warped and cropped Master-Print is passed to the second stage of the synthesis process where it is rendered with realistic textural details at 500 ppi. By passing different identity noise $z_{ID}$, distortion noise $z_{distort}$, and texture noise ($z_{texture}$), PrintsGAN is able to generate many fingerprint identities as well as impressions per identity. In this manner, PrintsGAN models both the inter-class and intra-class variance of a large fingerprint database, making it useful for training deep networks to extract representations for matching.}
\label{fig:schematic}
\end{figure*}

\section{Related Work}

Many studies have been conducted over the past several decades in an attempt to generate realistic synthetic fingerprints to address the paucity of publicly available fingerprint datasets. These approaches can be broadly categorized into i) ``hand-crafted" or  engineered approaches~\cite{cappelli2002synthetic, zhao2012fingerprint, johnson2013texture}, and ii) learning-based approaches~\cite{bontrager2018deepmasterprints, finger-gan, attia2019fingerprint, synfi, lightweight, kai_synthetic, mistry2019fingerprint, level-3, bahmani2021high}. 

While these approaches certainly made seminal contributions and tremendous strides towards realistic synthetic fingerprint datasets, they are also limited in a number of different ways. Qualitatively speaking, most of the existing synthetic fingerprint generators are not capable of generating fingerprints which are visually indistinguishable from real fingerprints. This can be seen by comparing the real fingerprints shown in Figure~\ref{fig:real_fingerprints} with the various synthetic fingerprints in Figure~\ref{fig:comparison}. This domain gap between real fingerprints and synthetic fingerprints renders the synthetic fingerprints of limited utility for both training deep networks and evaluation of fingerprint recognition systems.

Many of the ``hand-crafted" approaches are also limited via certain assumptions or restrictions imparted via the model chosen. For example:

\begin{itemize}
\item The models used to generate orientation fields (Zero Pole~\cite{sherlock1993model}), ridge-structure (AM/FM models~\cite{larkin2007coherent} or Gabor Filters~\cite{fogel1989gabor}) and minutiae points are assumed to be independent, creating unrealistic friction ridge patterns.

\item Fixed fingerprint ridge widths are often assumed. However, real fingerprints have varying ridge widths. In fact, the authors in~\cite{chen2014svm} showed that ridge width could be used to almost perfectly classify between real and synthetic fingerprints.

\item Common local minutiae configurations are not modeled, again enabling classification between real vs. synthetic fingerprints~\cite{gottschlich2014separating}.

\end{itemize}

More recent approaches to fingerprint synthesis aim to alleviate the shortcomings of some of the ``handcrafted" approaches by utilizing Generative Adversarial Networks (GANs) to learn the mapping from random noise to synthetic fingerprints without introducing some of the aforementioned assumptions. This has significantly improved the realism of synthetic fingerprints (Figure~\ref{fig:comparison}), however, it has introduced new limitations including:

\begin{itemize}
\item Many GAN based approaches focus on synthesizing small patches of fingerprints rather than full fingerprints to stabilize the training of the GAN.

\item The GANs are only capable of generating unique fingerprints. None of the existing GAN methods can generate multiple, full fingerprint impressions for a given fingerprint or model the intra-class variations.

\item A lack of training data results in some of the GAN based methods producing fingerprints which are even more dissimilar from real fingerprints than the `hand-crafted" approaches are capable of synthesizing.

\item GANs are naively utilized off-the-shelf without consideration of any fingerprint domain knowledge which can aid in improving the realism of the synthetic fingerprints. 

\end{itemize}

Like previous learning based synthesis methods~\cite{bontrager2018deepmasterprints, finger-gan, attia2019fingerprint, synfi, lightweight, kai_synthetic, mistry2019fingerprint, level-3, bahmani2021high}, PrintsGAN also utilizes several GANs to generate synthetic fingerprints which are more realistic than their handcrafted counterparts (Figure~\ref{fig:schematic}). However, PrintsGAN makes several key changes to the existing learning based synthesis pipeline in order to rectify their shortcomings. First, PrintsGAN utilizes domain knowledge during the synthesis process in a manner in which existing GAN based methods do not. Rather than naively learning a mapping directly from a random noise to a fingerprint via a single GAN, PrintsGAN breaks the synthesis process out into a series of steps each of which aims to model either inter-class variations or intra class variations. In particular, PrintsGAN uses one GAN $G_I(z)$ to generate a Master-Print (similar to the Master-Print generation used in hand-crafted approaches~\cite{cappelli2002synthetic}). Next PrintsGAN generates a non-linear warping and cropping of the Master-Print via a GAN $D_W(E_W(I))$ to simulate the effects of pressing a finger against a fingerprint reader platen at different roll, pitch, and yaw. Finally, PrintsGAN adds textural details to the warped and cropped Master-Print via a GAN $R_D(R_E(I))$. 

By using GANs to synthesize fingerprints, we leverage their ability to generate more realistic fingerprints than existing handcrafted methods. However, by utilizing domain knowledge from existing hand-crafted approaches via a series of synthesis steps that begin with a Master-Print, we are able to impart additional realism to our method and also the ability to control our GAN to generate multiple impressions for a given finger (something existing GAN-based fingerprint synthesis methods cannot do). In short, PrintsGAN aims to leverage the advantages of both the existing hand-crafted approaches as well as the more recent learning based approaches in order to address the limitations that both of them currently face when not synergistically tied together into a single algorithm like PrintsGAN.

As an addendum, we note that a plethora of work has been conducted in the realm of face recognition to generate synthetic faces via GANs, e.g.,~\cite{shoshan2021gan, karras2017progressive, karras2019style}. These methods generate highly realistic face images. Furthermore, the authors in~\cite{shoshan2021gan} impart explicit control over various facial attributes into the synthesis algorithm. Thus high quality faces and intra-class variations of the same face can be generated. However, such high quality synthetic full-image fingerprint GANs (with the ability to model intra-class variations) have not yet been proposed. We also note that work has been conducted to train face recognition models on synthetic face data~\cite{qiu2021synface, kortylewski2018training, kortylewski2019analyzing}. To the best of our knowledge, no such work has yet been conducted successfully within the field of fingerprint recognition where the motivation of such experimentation is much higher given the dearth of real fingerprint data in comparison to face data.

\begin{figure*}[t]
\includegraphics[scale=3.0]{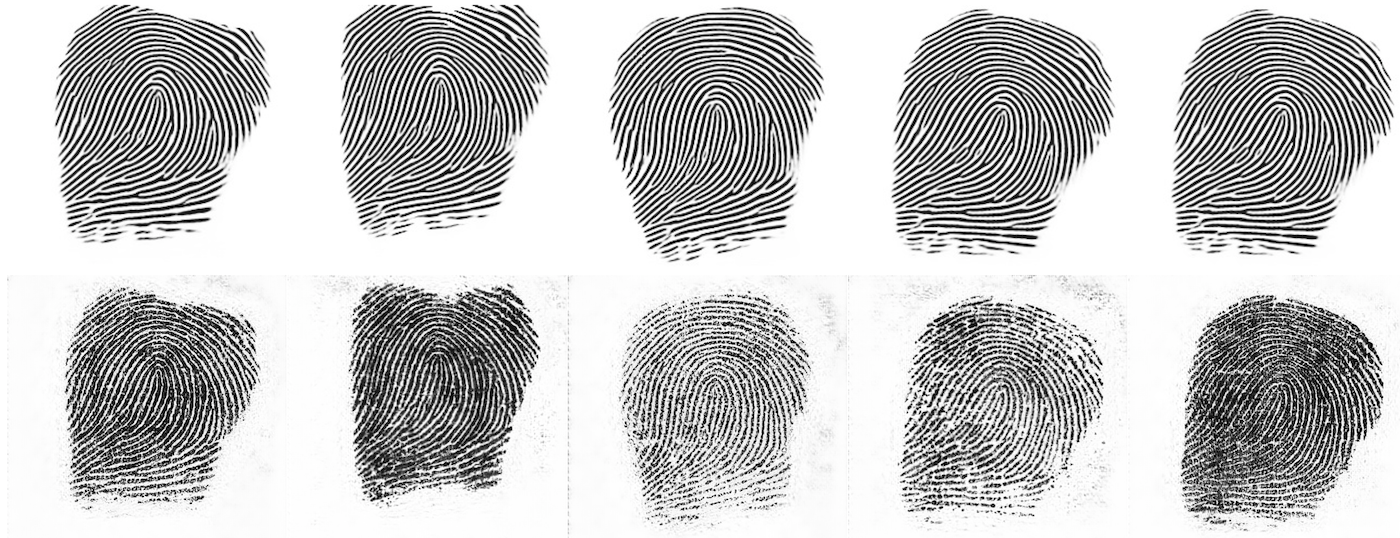} 
\caption{An example of a synthetic fingerprint identity, with five different impressions, generated by PrintsGAN. The top row shows the binary Master-Print with various warpings and croppings. The bottom row shows each of those Master-Print warps after a textural rendering.}
\label{fig:multiple_impressions}
\end{figure*}

\section{Approach}

PrintsGAN synthesizes fingerprints via a series of steps. First, a binary Master-Print $I_{ID}\in~\{0, 1\}^{256\times256}$ is generated via a random noise vector $z_{ID}\in~\mathbb{R}^{512}$, where $z$ is drawn from a continuous uniform distribution \textit{U(0, 1)} to create a new fingerprint identity. An example Master-Print can be seen in Figure~\ref{fig:schematic}. Next, $I_{ID}$ along with a warping noise vector $z_{distort}\in~\mathbb{R}^{16}$ is passed to a non-linear TPS warping module and cropping GAN $D_W(E_W(I_{ID}))$ to produce a warped Master-Print $I_w$. Finally, $I_w$ is passed to a renderer $R_D(R_E(I_w))$ along with a texture noise vector $z_{texture}~\in~\mathbb{R}^{128}$ to impart textural details to the final fingerprint $I_r$. Thus, by selecting different $z_{ID}$, we can generate many unique fingerprints. Likewise, by fixing $z_{ID}$, and selecting different $z_{distort}$ and $z_{texture}$, we can generate different impressions of the same fingerprint. Each of these steps are elaborated upon in the subsections below.

\subsection{Master-Print Synthesis}

The first step in the synthesis process requires learning a mapping from $z_{ID}\in~\mathbb{R}^{512}$ to a binary Master-Print $I_{ID}\in~\{0, 1\}^{256\times256}$. To perform this mapping, we utilize the BigGAN architecture~\cite{brock2018large} due to its demonstrated ability to produce a large variety of images (we want our fingerprint identities to be unique). The goal of the BigGAN generator is to generate a synthetic binary fingerprint. The discriminator must then try to distinguish between the synthetic binary fingerprint and a real binary fingerprint taken from an operational fingerprint database. More formally, the GAN is trained in accordance with the classic adversarial loss:

\begin{equation}
\label{eq:gan}
\mathcal{L}_{adv}(G_I, D) =  \mathbb{E}_x~[logD(x)] + \mathbb{E}_z~[log(1-D(G_I(z)))]
\end{equation} where $x$ is a binary fingerprint extracted from a real fingerprint. 

\begin{figure}[!t]
\includegraphics[scale=2.5]{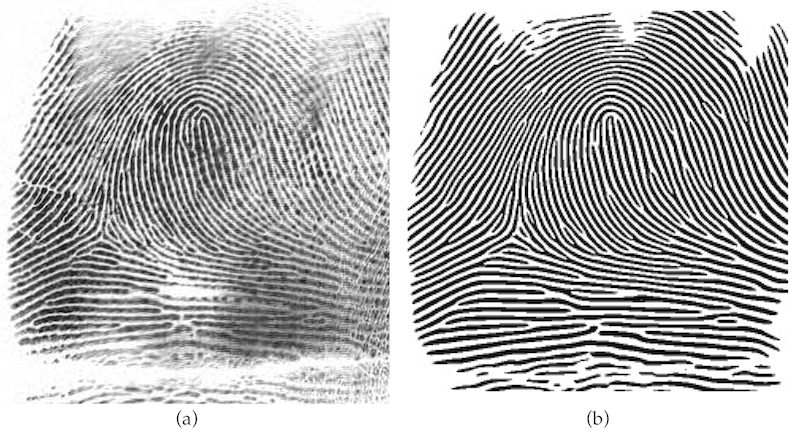} 
\caption{A rolled fingerprint from~\cite{longitudinal} is binarized via our trained grayscale fingerprint-to-binary auto-encoder.}
\label{fig:binary_image}
\end{figure}

\begin{figure*}[t]
\includegraphics[scale=3.0]{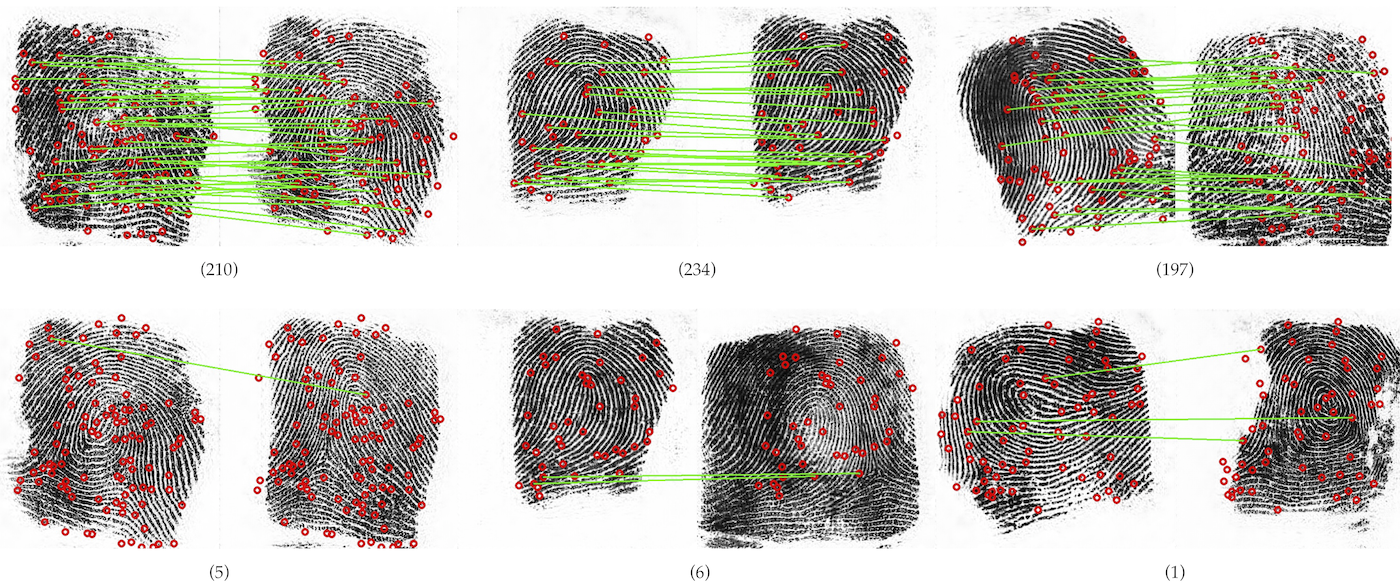} 
\caption{Examples of genuine pairs (top row) and imposter pairs (bottom row) synthesized by PrintsGAN. The minutiae matching score of Verifinger v12 SDK are displayed (below each pair) to show that i) PrintsGAN can generate unique fingerprints (low imposter pair scores) and ii) PrintsGAN can generate multiple impressions per finger (high genuine pair scores). Note, the matching threshold for Verifinger v12 for a False Acceptance Rate of $0.01\%$ is a match score of 48.}
\label{fig:gen_and_imp_matches}
\end{figure*}

For training the generator $G_{I}$ and the discriminator $D$ in Equation~\ref{eq:gan}, we utilize 282K unique fingerprints taken from the MSP longitudinal database used in~\cite{longitudinal} and~\cite{deepprint}. Prior to training, we extract binary fingerprint images from each of the 282K ``raw (grayscale) fingerprint images". To do this, we utilized a commercial fingerprint SDK (Verifinger v12 SDK) to first extract binary images from a subset of 10K raw fingerprint images. Then, we train an auto-encoder to learn the mapping from a raw fingerprint to a binary fingerprint using these 10K ground-truth binary fingerprints. More formally, given a raw fingerprint $I_{raw}$, we use an auto-encoder $R(.)$ to learn a mapping from $I_{raw}$ to a ground-truth binarized fingerprint $I_{binary}$ via an L-2 loss function\footnote{We also experimented with using a cross-entropy loss for this task since the output is a 0, 1 image, however, in practice, we found that the L-2 loss converged much more quickly and smoothly.}:

\begin{equation}
\label{eq:recon_loss}
\mathcal{L}_{recon} =  |R(I_{raw}) - I_{binary}|_2^2
\end{equation}

We note that we could directly use the commercial SDK to extract binary images from all 282K raw images, however, we specifically train $R(.)$ for this task for the following reasons. First, the commercial SDK is relatively slow, whereas $R$ enables us to quickly extract binary images for the full 282K database rather quickly, but more importantly, $R(.)$ is a differentiable binarization method and we intend to use it later on in a subsequent step as part of a loss function. An example of the binarization of $R(.)$ can be seen in Figure~\ref{fig:binary_image}.

\subsection{Warping and Cropping}

After training $G_{ID}$, we are able to generate binary Master-Prints. Each generated Master-Print comprises a new identity. The next step after generating each Master-Print is to impart a non-linear distortion and cropping to it to simulate the effects of placing a finger at different positions and pressures on a platen. For this step, we again utilize a GAN $G_w$ comprised of a content encoder $E_w$, a decoder $D_w$, and a warping encoder $L_w$. The content encoder encodes the Master-Print $I_{ID}$ into feature maps, while the warping encoder encodes a warping noise vector $z_{distort}\in~\mathbb{R}^{16}$ into a set of warping parameters $\Theta$. These warping parameters are then used to compute a Thin Plate Spline (TPS) warping transformation via $\mathcal{F}(I; \Theta)$. The decoder $D_w$ finally computes a segmentation mask $S\in~\{0, 1\}^{256\times256}$. The warped Master-Print is then computed via $\mathcal{F}(I; \Theta)\cdot S$.

Since the TPS warping module is differentiable, this entire process is trained via an adversarial loss (Equation~\ref{eq:gan}) where $x$ is a pair of real binary images from different impression of the same finger. Thus for the generator to fool the discriminator, it must generate realistic TPS distortions and croppings of the input Master-Prints to mimic the distortions and croppings between a pair of real binary fingerprints derived from the same finger. Examples of different warpings and croppings of a Master-Print are shown in the top row of Figure~\ref{fig:multiple_impressions}.

\subsection{Renderer}

Finally, after warping and cropping different portions of a Master-Print into $I_w$, we pass it through one final GAN, $G_r$ to add realistic textural details. For the GAN architecture we again utilize the BigGAN\footnote{We also experimented with StyleGAN~\cite{karras2019style} for the renderer, but found it more difficult to maintain the identity.} architecture due to the realism it imparts to high resolution images~\cite{brock2018large}. However, in order to impart different textural details to different impressions of the same finger, we add a texture encoder to the BigGAN architecture. In particular, we add as input, a texture noise vector $z_{texture}\in~\mathbb{R}^{128}$. This noise vector is then encoded to a $\gamma$ and $\beta$ for performing instance normalization on the BigGAN feature maps~\cite{ulyanov2016instance}. As was shown in~\cite{ulyanov2016instance}, instance normalization can be used in GANs to modulate between different styles in GAN images. In our case, modulating these styles results in different types of impression noise between each impression of an input fingerprint. To ensure that we maintain identity after rendering, we pass our rendered images, $I_r$ through the trained binarization auto-encoder $R$. We then make sure that the binary image of the rendered image is the same as the original input binary image, $I_w$, via an L2 loss, i.e., we minimize $|R(I_r) - I_w|_2^2$. This ensures that we do not introduce any new friction ridge patterns during the rendering step. To ensure that the rendered fingerprint looks realistic, we again utilize an adversarial loss (Eq.~\ref{eq:gan}). Examples of these different textural renderings of an input Master-Print can be seen in the bottom row of Figure~\ref{fig:multiple_impressions}.

\section{Experimental Results}

\subsection{Qualitative Results}
Qualitatively, the fingerprint impressions generated by PrintsGAN are both highly realistic and discriminative. As seen in the bottom row of Figure~\ref{fig:gen_and_imp_matches}, PrintsGAN is able to generate unique fingerprints that yield low imposter scores with a SOTA commercial fingerprint matcher, Verifinger v12. Besides generating high quality impressions of multiple fingerprint identities, PrintsGAN is also able to simulate realistic intra-class variation between impressions of the same fingerprint identity via its texture and style rendering stage, while achieving high similarity between genuine impressions of the same finger (top row of Figure~\ref{fig:gen_and_imp_matches}). 

\begin{figure*}
\begin{center}
\includegraphics[width=0.95\linewidth]{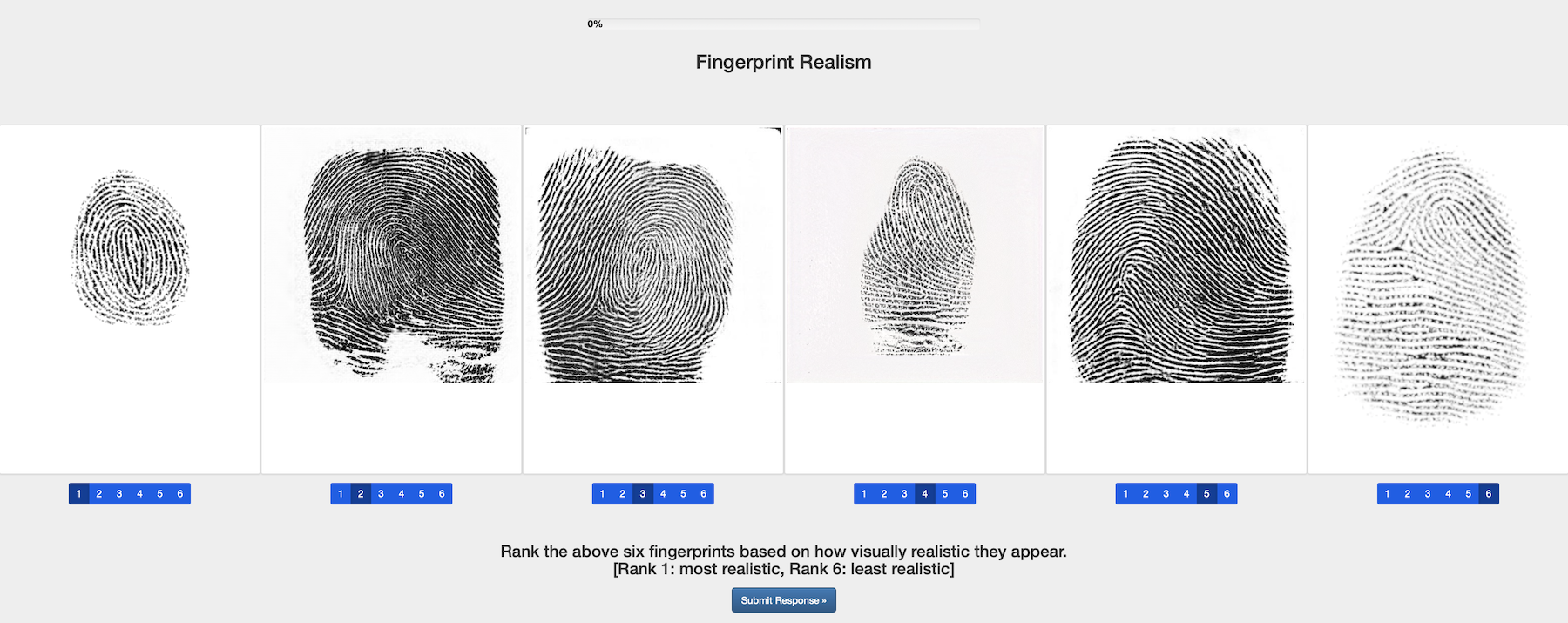} 
\caption{Example prompt from experiment 1 of the expert survey on synthetic fingerprint realism. Participants were asked to rank the six synthesis methods in order of realism on a scale from 1-6, 1 being the most realistic to 6 being the least realistic.}
\label{fig:survey1}
\end{center}
\end{figure*}

\subsection{CrowdSource Evaluation}
To validate the realism of the fingerprint images generated by PrintsGAN, we performed a crowdsourcing experiment where sixteen fingerprint domain researchers visually assessed the quality of our synthetic fingerprints. In particular, we designed experiments to test 1) how our fingerprints look with respect to existing baseline methods and 2) how well PrintsGAN fingerprints mimic real fingerprints.

For experiment 1, experts were presented with images from six different synthetic fingerprint generators (~\cite{cappelli2002synthetic, bahmani2021high, mistry2019fingerprint, kai_synthetic, novetta}, and PrintsGAN) and asked to order each fingerprint from most realistic (1) to least realistic (6) on a scale from 1-6\footnote{These 6 methods were chosen because of their visual realism and ability to synthesize full fingerprint impressions. Note, some methods, e.g., L3-SF~\cite{level-3}, despite being very realistic, were not chosen because they do not generate full fingerprint images.}. The experiment consisted of 15 different trials which consisted of a random selection of images from each method presented in a randomly assigned order where each trial was kept the same for each expert. An example of one of these trials is given in Figure~\ref{fig:survey1}. The mean and standard deviation of the ratings for each method are presented in Table~\ref{tab:survey1_results}. Out of the three rolled fingerprint synthesis methods, PrintsGAN obtains the best average realism rating of $2.45 \pm 1.47$, which is a significant improvement compared to the next best method by Mistry et al.~\cite{mistry2019fingerprint} that attained a score of $3.70 \pm 1.61$. For the plain print methods, CFG~\cite{bahmani2021high} scores the best overall ranking of $2.27 \pm 1.41$; however, unlike PrintsGAN, CFG is unable to generate impressions of specific identities, which limits its utility in both training and evaluating fingerprint recognition algorithms. Furthermore, it should also be noted that generating a rolled fingerprint is more challenging than plain fingerprints because of its larger surface area, distortion and number of minutiae points.

\begin{figure}
\centering
\includegraphics[width=0.8\linewidth]{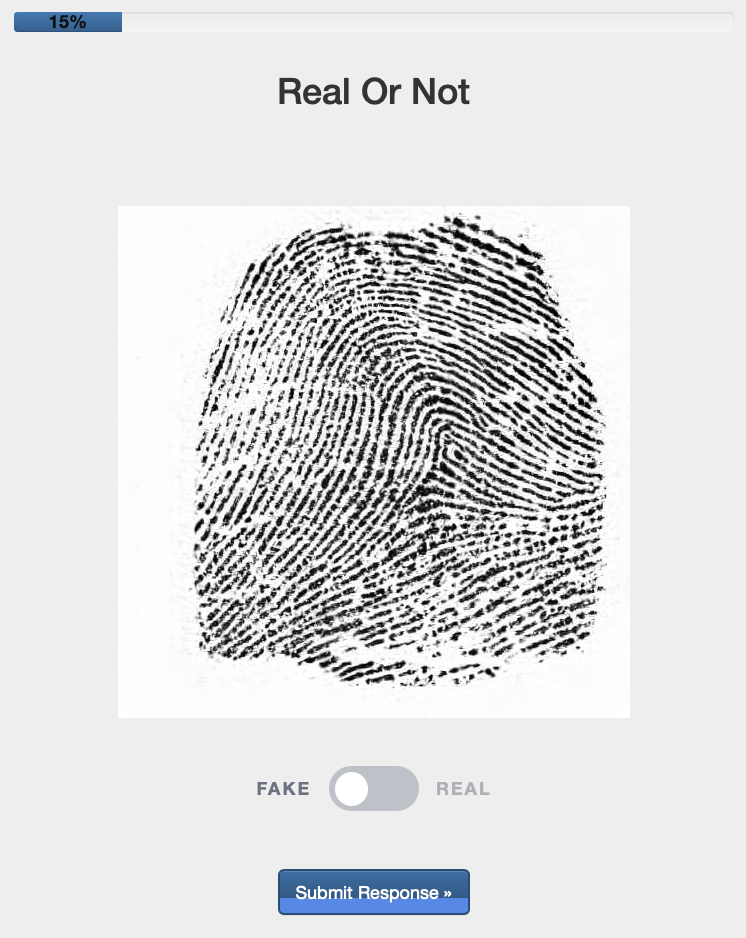} 
\caption{Example prompt for experiment 2 of the expert survey on synthetic fingerprint realism. Experts were asked to chose between real or fake when presented with a fingerprint image.}
\label{fig:survey2}
\end{figure}

For experiment 2, participants were presented (one at a time) a total of $130$ synthetic and real fingerprint images and asked whether each individual fingerprint was real or fake. An example prompt for one of these $130$ trials is presented in Figure~\ref{fig:survey2}. An interesting observation from this study is that the standard deviation among plain methods (both synthetic and real) is typically much larger compared to the rolled fingerprint methods. This seems to suggest that it is more difficult for experts to distinguish between synthetic and real plain prints compared to rolled prints. Furthermore, CFG was the most successful in passing as real
among plain print methods, achieving an impressive $84.07 \pm 21.14\%$ classification rate as real. In comparison, that is surprisingly higher than the classification rate of the real plain fingerprint database sampled from the FVC 2002 DB1-A dataset~\cite{maio2002fvc2002}. Similarly, PrintsGAN was the most successful of the rolled fingerprint methods, achieving a $66.67 \pm 20.22\%$ real classification rate, which is much closer to the real rolled database (sampled from \cite{longitudinal}) classification rate of $72.04 \pm 18.81\%$ compared to the baseline methods.

\begin{table}
\caption{Results for Expert Crowdsourcing Experiment 1: Rate Each Fingerprint from Most Realistic to Least Realistic (1 = most realistic, 6 = least realistic).}
\begin{tabular}{|l|c|c|c|}
\hline
                       & \textbf{Type} & \textbf{Average Realism Rating} & \textbf{Std. Dev} \\ \hline
\textit{Sfinge~\cite{cappelli2002synthetic}}        & Plain         & 3.67                                      & 1.61              \\ \hline
\textit{IBG Novetta~\cite{novetta}}   & Plain         & 3.19                                      & 1.48              \\ \hline
\textit{CFG~\cite{bahmani2021high}}          & Plain         & \textbf{2.27}                                      & \textbf{1.41}              \\ \noalign{\hrule height 1.0pt}
\textit{Kai et al.~\cite{kai_synthetic}}    & Rolled        & 4.33                                      & 1.43              \\ \hline
\textit{Mistry et al.~\cite{mistry2019fingerprint}} & Rolled        & 3.70                                      & 1.61              \\ \hline
\textit{PrintsGAN}     & Rolled        & \textbf{2.45}                                      & \textbf{1.47}              \\ \hline
\end{tabular}
\label{tab:survey1_results}
\end{table}

\begin{table}
\caption{Results for Expert Crowdsourcing Experiment 2: Rate Each Fingerprints as Real or Fake.}
\begin{tabular}{|lccc|}
\hline
\multicolumn{4}{|c|}{\textbf{Percentage Classified as Real per Dataset (Mean $\pm$ Std. Dev.)}}                                                                                                                                                          \\ \hline
\multicolumn{2}{|c|}{\textbf{Plain Prints}}                                                                               & \multicolumn{2}{c|}{\textbf{Rolled Prints}}                                                           \\ \hline
\multicolumn{1}{|l}{\textit{Sfinge~\cite{cappelli2002synthetic}}}      & \multicolumn{1}{c|}{37.74 $\pm$ 39.13}            & \multicolumn{1}{c}{\textit{Kai et al.~\cite{kai_synthetic}}}    & \multicolumn{1}{c|}{19.58 $\pm$ 21.70}            \\
\multicolumn{1}{|l}{\textit{IBG Novetta~\cite{novetta}}} & \multicolumn{1}{c|}{40.51 $\pm$ 30.62}            & \multicolumn{1}{c}{\textit{Mistry et al.~\cite{mistry2019fingerprint}}} & \multicolumn{1}{c|}{35.83 $\pm$ 22.56}            \\
\multicolumn{1}{|l}{\textit{CFG~\cite{bahmani2021high}}}        & \multicolumn{1}{c|}{\textbf{84.07 $\pm$ 21.14}}            & \multicolumn{1}{c}{\textit{PrintsGAN}}     & \multicolumn{1}{c|}{\textbf{66.67 $\pm$ 20.22}}            \\ \hline
\multicolumn{1}{|l}{\textit{Real Plain}}  & \multicolumn{1}{c|}{67.50 $\pm$ 36.61}            & \multicolumn{1}{c}{\textit{Real Rolled}}   & \multicolumn{1}{c|}{72.04 $\pm$ 18.81}            \\ \hline
\end{tabular}
\label{tab:survey2_results}
\end{table}

\begin{table*}[t!]
\centering
\caption{\normalsize{Fingerprint metrics for real (DB-1) and PrintsGAN (DB-2) fingerprints. Minutiae quality and NFIQ2 scores both have a range of [0, 100].}}
\begin{tabularx}{\textwidth}{l|YY|YY}
\noalign{\hrule height 1.5pt}
                                  & \multicolumn{2}{c|}{DB-1 (Real)}                          & \multicolumn{2}{c}{DB-2 (PrintsGAN)}                     \\
Measure                           & \multicolumn{1}{c}{Mean} & \multicolumn{1}{c|}{Std. Dev.} & \multicolumn{1}{c}{Mean} & \multicolumn{1}{c}{Std. Dev.} \\
\noalign{\hrule height 1.0pt}
Total Minutiae Count              & 92.70                    & 24.16                         & 79.30                    & 16.59                         \\
Ridge Ending Minutiae Count       & 49.98                    & 16.06                         & 43.00                    & 10.14                         \\
Ridge Bifurcation Minutiae Count  & 42.71                    & 13.36                         & 36.30                    & 9.37                          \\
Quality of Minutiae               & 73.50                    & 15.07                         & 72.14                    & 16.05                         \\
Fingerprint Area (Megapixels) & 0.179               & 0.038                      & 0.171               & 0.019                      \\
Fingerprint Image Quality (NFIQ2)                             & 54.21                    & 22.73                         & 63.63                    & 21.38    \\  
\noalign{\hrule height 1.5pt}
\end{tabularx}
\label{tab:fp_stats}
\end{table*}

\subsection{Quantitative Evaluation}
In line with previous research on synthetic fingerprint generation, we have chosen to evaluate PrintsGAN in terms of several quantitative metrics, including the distribution of minutiae count, type, and quality extracted from PrintsGAN generated fingerprints compared to the real fingerprint training set, the distribution of genuine and imposter match scores from two SOTA fingerprint matchers, NFIQ2 quality scores, and closed-set identification experiments. In particular, let us denote the first database of $282$k real rolled fingerprints images from the training database~\cite{longitudinal} as DB-1 and the database of synthetic rolled fingerprints from PrintsGAN as DB-2. DB-2 consists of $35,000$ synthetic fingers, each with $15$ impressions per finger, totaling $525,000$ fingerprint images. Let DB-3 denote a third real, rolled fingerprint database from NIST Special Database 4~\cite{sd4} as an additional reference database to compare with the PrintsGAN generated database. DB-3 contains $2,000$ unique fingerprints with $2$ impressions per finger.

\subsubsection{Fingerprint Metrics}

First, we have computed some fingerprint metrics from the distribution of real fingerprints (DB-1) and synthetic fingerprints from PrintsGAN (DB-2). The metrics regarding minutiae (total number of minutiae, number of minutiae ridge endings, number of minutiae bifurcations, and minutiae quality) were computed using the Verifinger v12 SDK and the mean and standard deviation for each are given in Table~\ref{tab:fp_stats}. Additionally, the average fingerprint area and NFIQ2 quality scores were computed for each database. PrintsGAN is able to generate a diverse set of fingerprint identities, which is supported by the large standard deviation across all these metrics. It is also able to generate fingerprints which resemble the real, rolled training database in terms of fingerprint area, minutiae statistics, and NFIQ2 scores; albeit, the number of minutiae detected in PrintsGAN fingerprints is, on average, slightly less compared to the real fingerprint database. This is in part due to the slightly smaller fingerprint area generated by PrintsGAN along with the higher average NFIQ2 scores, which results in less spurious minutiae. Quantitatively, PrintsGAN fingerprints give a higher Goodness Index (GI)~\cite{ratha1995adaptive} compared to the real fingerprint database ($0.00058$ vs $-0.00084$), which can be attributed to the decrease in spurious minutiae.

\subsubsection{Imposter Distributions}

Next, we have computed several distributions of imposter scores to 1) detect identity leakage from the real fingerprint training database in the synthetic fingerprints generated from PrintsGAN and 2) evaluate the uniqueness of fingerprint identities being generated by PrintsGAN. To detect leakage in the generated fingerprints, we have computed match scores between each of the $282$k training fingerprint identities in real DB-1 to each unique fingerprint identity in PrintsGAN DB-2. In total, the number of comparisons would be $148$ billion ($35,000\times 15\times 282000$), instead we randomly select one impression per the $35,000$ fingers from DB-2 and compute $9.87$ billion match scores. The match scores are computed in a 2-step process in line with the 2-stage search procedure implemented in \cite{deepprint} to significantly reduce the computational time required to perform $9.87$ billion matches. In the first stage, DeepPrint~\cite{deepprint} is used to filter out the matches which obtained a match score lower than $0.83$ (these are obvious non-matches), a threshold which was empirically obtained for a $0.01\%$ False Acceptance Rate on the NIST Special Database 4~\cite{sd4}. This stage yielded $32,182$ pairs which were then subjected to the Verifinger v12 ISO minutiae-matcher operating at a match threshold of $48$ set by Neurotechnology for a FAR of $0.01\%$. As a result, out of the $35,000$ unique fingers in DB-2, only a mere $15$ ($0.04\%$ of the database) had Verifinger match scores above $48$ with any of the $282$k unique fingers in the real training database, DB-1. Furthermore, each of these $15$ fingers just barely exceeded the threshold of $48$ for genuine matches, of which the maximum score recorded was a score of $65$. Even so, we acknowledge that this is still \textit{some} degree of risk and information leakage, and so we have removed these images from the training set before releasing it to the public.

\begin{figure}
\centering
\subfloat[]{\includegraphics[width=0.95\linewidth]{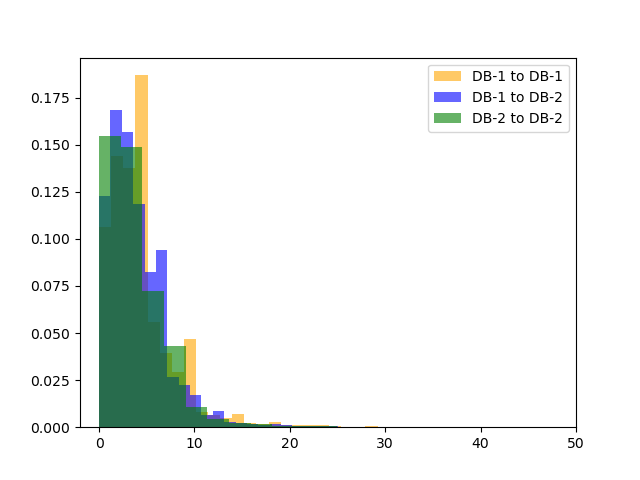}
\label{hists}}\\
\subfloat[]{\includegraphics[width=0.95\linewidth]{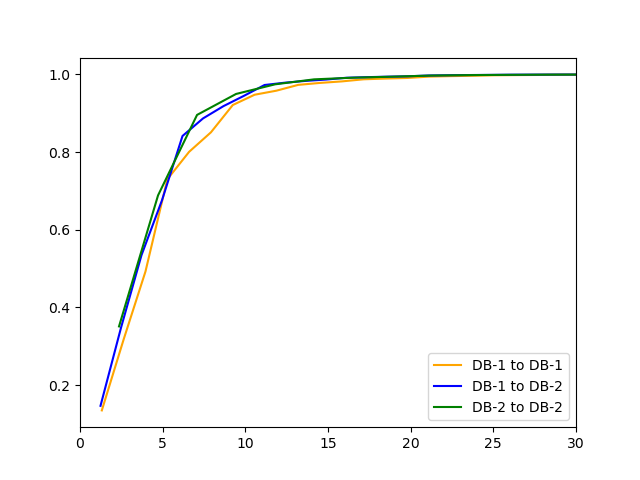}
\label{cdfs}}
\caption{Histograms (a) and CDFs (b) for imposter score distributions.}
\label{fig:imposter_plots}
\end{figure}

To investigate the uniqueness of fingerprint identities being generated by PrintsGAN, we have computed several distributions of imposter scores: i) imposter scores within DB-1, ii) imposter scores within DB-2, and iii) imposter scores between DB-1 v. DB-2. In particular, we have randomly sampled $35$k unique fingerprints from DB-1 and one impression from each of the $35$k unique fingerprints in DB-2 to perform the comparisons. Within each histogram, there are a total of $1.225$ billion imposter matches computed, from which we randomly sample $35$ million scores. Figure~\ref{fig:imposter_plots} shows the histograms as well as the Cumulative Distribution Function (CDF) for each of the three histograms.

Next, we computed the non-parametric independent samples Kolmogorov-Smirnov (KS) test between the empirical imposter distributions to confirm the distinctness of the synthetic fingerprints, similar to \cite{bahmani2021high}. The KS test is computed between the empirical imposter distribution of DB-1 and the empirical imposter distribution between DB-2 v. DB-1 with an alternative hypothesis that the imposter distribution between DB-2 v. DB-1 is greater than the distribution within DB-1. Note, that the alternative hypothesis describes the CDFs of the imposter distributions; thus, suppose if $x_1 ~ F$ and $x_2 ~ G$, then $F(x)$ is greater than $G(x)$ would indicate that the values in $x_1$ tend to be less than those in $x_2$. In this case, our experiment yielded a KS statistic of $0.0462$ and a p-value of $0.0$, which indicates rejecting the null hypothesis in favor of the alternative hypothesis. Since the distribution of DB-2 v. DB-1 imposter scores is greater than the distribution of imposter scores within DB-1, we conclude that the imposter scores between DB-2 v DB-1 tend to be less than the imposter scores within DB-1. Thus, the synthetic fingerprints generated by the PrintsGAN are distinct from real samples in DB-1.

\begin{figure}
\centering
\includegraphics[width=1.0\linewidth]{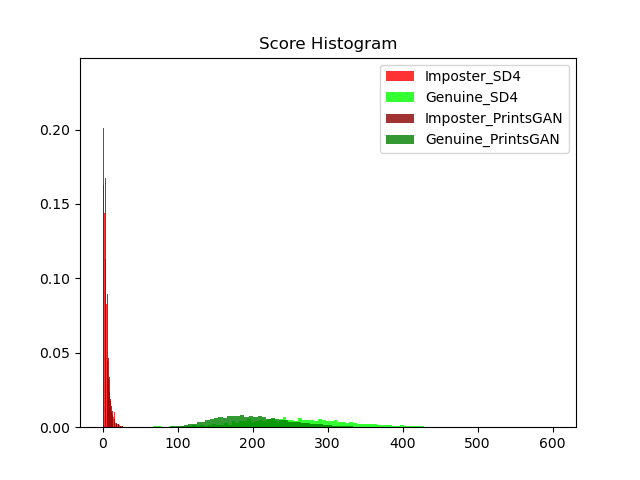}
\caption{Match score distributions for DB-1 (real) and DB-2 (PrintsGAN), computed with the Verifinger v12 SDK.}
\label{fig:verifinger_scores}
\end{figure}

To further show that the distribution of fingerprints generated by PrintsGAN follow what we would expect from a distribution of real, rolled fingerprints, we have computed genuine and imposter matches with Verifinger for both DB-3 and DB-2. The score histograms are shown in Figure~\ref{fig:verifinger_scores}, showing similar distributions for both DB-2 and DB-3 as well as excellent separation between mated and non-mated scores. The distribution of genuine scores for DB-2 is slightly shifted to the right compared to DB-3, which correlates with the slightly higher NFIQ2 scores computed for DB-2 compared to DB-3. Thus, PrintsGAN tends to generate fewer samples of very poor NFIQ2 quality compared to what we might expect in a real, operational fingerprint dataset.

\begin{table*}
\normalsize{
\caption{Authentication Accuracy. The first two rows correspond to training a deep network with just synthetic prints, the third row corresponds to training on N2N alone, and the final two rows correspond to pretraining with synthetic and then finetuning on N2N~\cite{nist302}.}
\begin{tabularx}{\textwidth}{X|M|M|M}
\noalign{\hrule height 1.5pt}
\textbf{Dataset} & \textbf{NIST SD4~\cite{sd4}} & \textbf{FVC 2002 DB1 A~\cite{fvc2002}} & \textbf{FVC 2004 DB1 A~\cite{fvc2004}} \\
 & \textbf{TAR @ 0.01\% FAR} & \textbf{TAR @ 0.01\% FAR} & \textbf{TAR @ 0.01\% FAR} \\
\noalign{\hrule height 1.0pt}

Sfinge$^\ddagger$ & 10.78 \pm 0.88\% & 12.57 \pm 3.08\% & 20.80 \pm 0.48\% \\
\hline
PrintsGAN$^\ddagger$ & 52.65 \pm 2.33\% & 59.59 \pm 5.13\% & 22.35 \pm 4.89\% \\
\noalign{\hrule height 1.0pt}
N2N$^\dagger$ & 73.37 \pm 3.15\% & 79.68 \pm 3.67\% & 65.99 \pm 8.27\% \\
\hline
N2N$^\dagger$ + Sfinge$^\ddagger$ & 54.70 \pm 2.83\% & 56.68 \pm 7.42\% & 62.68 \pm 1.06\% \\
\hline
N2N$^\dagger$ + PrintsGAN$^\ddagger$ & \mathbf{87.03 \pm 0.33\%} & \mathbf{89.74 \pm 0.22\%} & \mathbf{90.22 \pm 1.19\%} \\
\noalign{\hrule height 1.0pt}
\multicolumn{4}{l}{$^\dagger$(2k IDs, 10 impressions), $^\ddagger$(35k IDs, 15 impressions)}\\

\end{tabularx}
\label{tab:dp_training}
}
\end{table*}

\subsection{Training Deep Networks}
A significant motivation in designing PrintsGAN to be able to synthesize a large-scale database of many fingerprint identities with sufficient number of impressions per identity is to facilitate both the training and evaluation of deep networks for fixed-length fingerprint representations. To the best of our knowledge, the two methods in the open academic literature with the capability of synthesizing multiple impressions of a particular fingerprint identity, L3-SF~\cite{level-3} and Sfinge~\cite{cappelli2002synthetic}, either only synthesize partial fingerprints or lack sufficient realism to be effective for this purpose. 

To benchmark the effectiveness of PrintsGAN compared to the existing methods in improving the training of deep network-based fingerprint representation algorithms, we have compared the performance of multiple DeepPrint models~\cite{deepprint} trained on PrintsGAN generated fingerprints and the baseline synthetic fingerprint method Sfinge in terms of True Acceptance Rate (TAR) on test databases of real fingerprints\footnote{We chose not to benchmark against L3-SF since L3-SF does not generate full fingerprint impressions.}. It is the goal that deep network models, like DeepPrint, pretrained on synthethic fingerprint datasets and finetuned on the publicly available, real fingerprint data, will outperform the same deep network model trained on only the limited set of publicly available, real fingerprint data.

One of the largest publicly available, real fingerprint training database is the NIST Special Database 302 (i.e., N2N Database), which contains $25$k total fingerprints from $2,000$ distinct fingers. As a benchmark for this experiment, we trained a DeepPrint model on the N2N database alone and evaluated its performance on three real fingerprint databases, NIST SD4~\cite{sd4}, FVC 2002 DB1-A~\cite{fvc2002}, and FVC 2004 DB1-A~\cite{fvc2004}. Recall that NIST SD4 is comprised of rolled fingerprints whereas FVC databases are comprised of plain fingerprints. Then, to observe the utility of incorporating synthetic fingerprint data to augment the training of deep network fingerprint recognition models, we performed a procedure of pretraining on synthetic data and finetuning on the real, NIST N2N data. The utility of the synthetic fingerprints is then measured in the performance improvement of these finetuned models over the model trained on real data alone.

For our experiments, we trained on equal amounts of synthetic fingerprints from both our method and from Sfinge ($35,000$ unique fingerprint identities with $15$ impressions each). The performance of these models trained on synthetic data, along with the model trained on NIST N2N only, are given in Table~\ref{tab:dp_training}. We trained each model three times and reported the mean and standard deviation of the performance across the three test datasets. We notice that the performance of the model trained only on 35k PrintsGAN identities exceeds that of the models trained on Sfinge data on all three datasets. More importantly, when finetuning the model pretrained on PrintsGAN data on the N2N dataset, the performance exceeds that of the model trained on N2N from scratch, demonstrating the utility of PrintsGAN synthetic fingerprints to boost the performance of deep network-based models beyond what is attainable training on the limited amount of real fingerprint datasets alone. Furthermore, the performance improvement on both rolled (NIST SD4) and plain print (FVC) datasets shows that the rolled fingerprints generated by PrintsGAN are even useful for improving the performance of DeepPrint on plain fingerprints.




\begin{figure}
\centering
\includegraphics[width=1.0\linewidth]{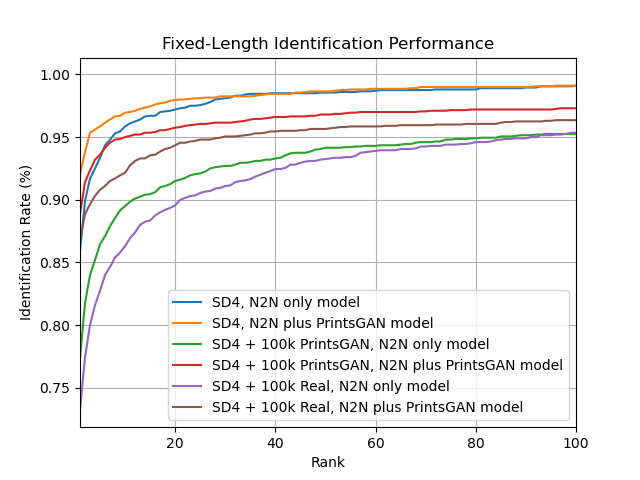} 
\caption{Closed-set identification accuracy of DeepPrint models trained on N2N data only vs. N2N + PrintsGAN data. The various curves shown are comparing the search performance of these two models on i.) SD4, ii.) SD4 augmented with 100k Real fingerprint images, and iii.) SD4 augmented with 100k PrintsGAN fingerprints. Best viewed in color.}
\label{fig:search}
\end{figure}

\subsection{Search Experiments}
In addition to improving the authentication performance of deep network models via augmenting the training datasets with synthetic fingerprints from PrintsGAN, the identification accuracy also shows a similar improvement - which happens to be an application which greatly benefits from fixed-length representations due to the drastically improved search speed. Figure~\ref{fig:search} compares the closed-set identification accuracy of a DeepPrint model trained on only N2N data vs. the DeepPrint model which was pretrained on PrintsGAN images and finetuned on N2N. The search results on SD4 show considerable improvement for the model trained on both N2N + PrintsGAN data, with a rank 1 identification rate of $92.05\%$ compared to $85.90\%$ for training on N2N alone. Further augmenting the gallery with $100$k real fingers from \cite{longitudinal} shows a similar improvement ($86.80\%$ vs. $73.20\%$). Instead of augmenting the gallery with $100$k real fingers, we can use $100$k fingers generated from PrintsGAN to augment the gallery. In doing so, the identification performance is similar to the performance obtained by augmenting the gallery with real fingerprints (comparing the green vs. purple curves in Figure~\ref{fig:search}), demonstrating that our synthetic prints generated by PrintsGAN can be used for benchmarking large-scale identification of fingerprint recognition systems.

\subsection{Computational Efficiency}
DeepPrint models and PrintsGAN training code are implemented in Tensorflow 1.14.0. All models were trained across 2 NVIDIA GeForce RTX 2080 Ti GPUs. Since PrintsGAN involves a multi-stage generation process, the time required to synthesize a fingerprint by PrintsGAN is an aggregation of the time required by each of the three stages: i.) master print synthesis (20.0ms), ii.) warping and cropping (36.5ms), and iii.) textural rendering (42.7ms). Thus, the total time required to synthesis one fingerprint is approximately 99.2ms. As a future improvement to speed up the generation time, these three stages could be condensed into a single network. 

Each fingerprint generated by PrintsGAN is a $512\times512$ 8-bit grayscale image requiring about 256KB of storage. Thus, storage requirements for such a large dataset may quickly become a concern. However, since PrintsGAN is deep network which is fully implemented in Tensorflow, instead of generating a large database beforehand, one could elect to generate fingerprint samples on the fly during the training of a deep network-based fingerprint recognition system (e.g., DeepPrint).

\section{Conclusion}
In this work, we developed a GAN-based fingerprint synthesis method, PrintsGAN, that is capable of generating high-quality, $512\times 512$ resolution, rolled fingerprints that closely resemble the minutiae quantity, type, and quality distributions of an operational, real fingerprint database on which the method was trained. Further experiments validate the capability of PrintsGAN to synthesize diverse and unique fingerprint identities with realistic intra-class variation. PrintsGAN is able to achieve both improved realism (supported by survey results from domain-experts in fingerprint biometrics) and control over the synthesis process through a synergy of traditional domain-knowledge synthesis methods and state-of-the-art GAN methods. Most importantly, the utility of PrintsGAN generated fingerprints was demonstrated through improved recognition performance (when training on synthetic data) beyond what was achievable for a similar deep network model architecture on the paucity of publicly available, real fingerprint data. Furthermore, experimental results show that PrintsGAN does not leak any significant information from the database of real training images, permitting us to publicly release a large-scale dataset of synthetically generated samples from PrintsGAN, without the privacy concerns that restrict the release of current databases of real fingerprints.


%

\ifCLASSOPTIONcompsoc
  \section*{Acknowledgments}
\else
  \section*{Acknowledgment}
\fi

The authors would like to thank the fingerprint researchers who participated in our crowdsourcing experiments to compare various synthetic fingerprint generators. This research was partially supported by a grant from the National Institute of Standards and Technology (NIST).

\ifCLASSOPTIONcaptionsoff
  \newpage
\fi



\bibliographystyle{IEEEtran}
\bibliography{cite}
%



%

\begin{IEEEbiography}[{\includegraphics[width=1in,height=1.25in,clip,keepaspectratio]{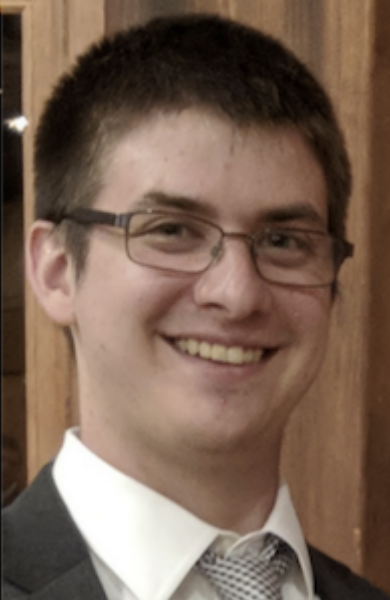}}]{Joshua J. Engelsma} graduated magna cum laude with a B.S. degree in computer science from Grand Valley State University, Allendale, Michigan, in 2016. He completed his PhD degree in Computer Science at Michigan State University in 2021. His research interests include pattern recognition, computer vision, and image processing with applications in biometrics. He won the best paper award at the 2019 IEEE International Conference on Biometrics (ICB), and the 2020 Michigan State University College of Engineering Fitch Beach Award.
\end{IEEEbiography}

\begin{IEEEbiography}[{\includegraphics[width=1in,height=1.25in,clip,keepaspectratio]{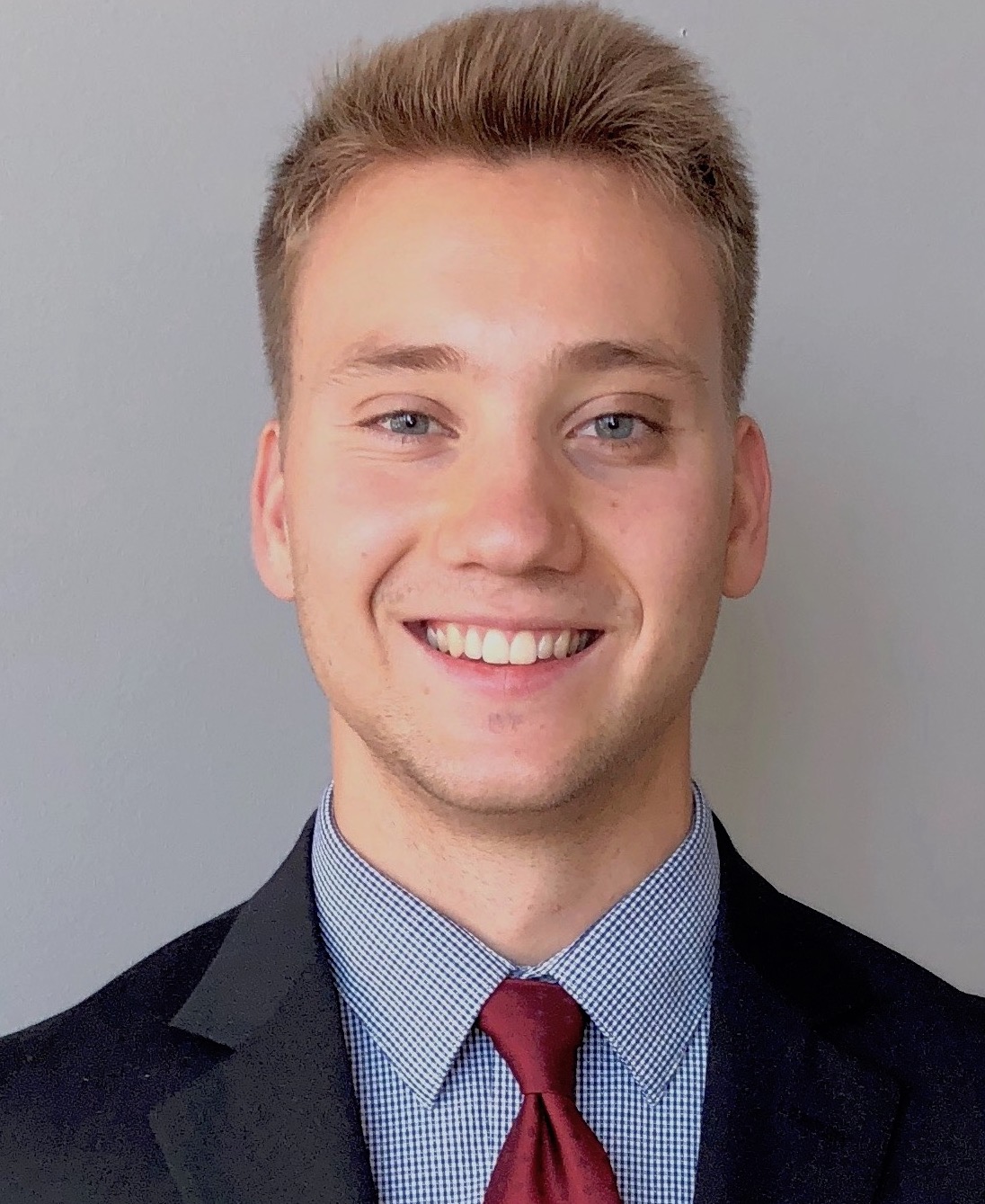}}]{Steven A. Grosz}
received his B.S. degree with highest honors in Electrical Engineering from Michigan State University, East Lansing, Michigan, in 2019. He is currently a doctoral student in the Department of Computer Science and Engineering at Michigan State University. His primary research interests are in the areas of machine learning and computer vision with applications in biometrics.
\end{IEEEbiography}

\begin{IEEEbiography}[{\includegraphics[width=1in,height=1.25in,clip,keepaspectratio]{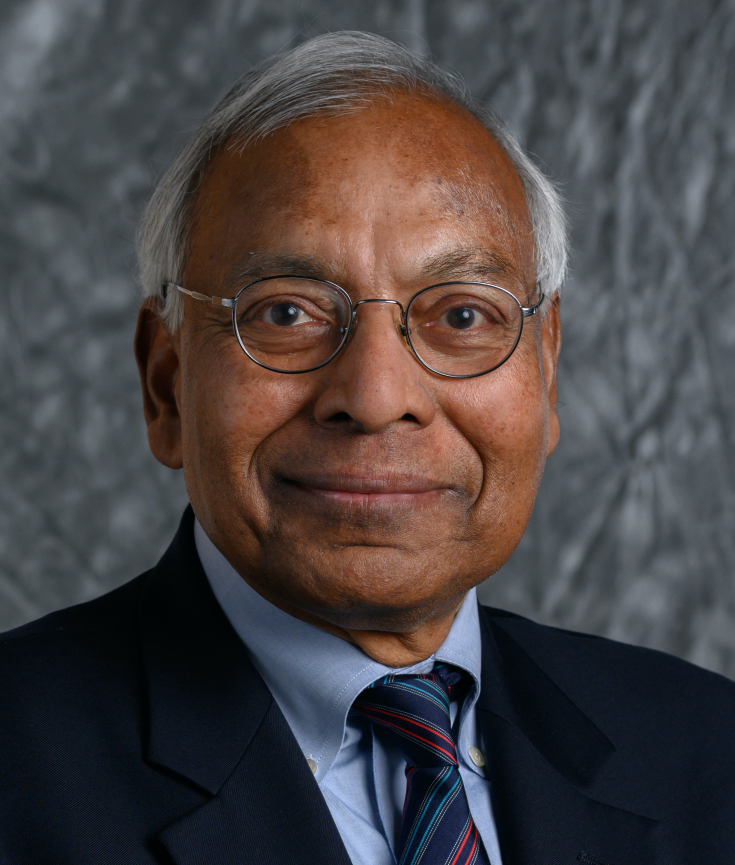}}]{Anil K. Jain}
Anil K. Jain is a University distinguished professor in the Department of Computer Science and Engineering at Michigan State University. His research interests include pattern recognition and biometric authentication. He served as the editor-in-chief of the IEEE Transactions on Pattern Analysis and Machine Intelligence and was a member of the United States Defense Science Board. He has received Fulbright, Guggenheim, Alexander von Humboldt, and IAPR King Sun Fu awards. He was elected to the National Academy of Engineering, the Indian National Academy of Engineering, the World Academy of Sciences, and the Chinese Academy of Sciences.
\end{IEEEbiography}





\end{document}